\documentclass[runningheads]{llncs}
\usepackage{graphicx}
\usepackage{float}
\usepackage{amsmath}
\usepackage{xcolor}

\begin{document}
\title{HydraMix-Net: A Deep Multi-task Semi-supervised Learning Approach for Cell Detection and Classification}
\titlerunning{HydraMix-Net}
\author{R.M. Saad Bashir\inst{1} \and
Talha Qaiser\inst{2} \and
Shan E Ahmed Raza\inst{1} \and
Nasir M. Rajpoot\inst{1,3}}

\authorrunning{R.M.S Bashir et al.}
\institute{Department of Computer Science, University of Warwick, Coventry, UK \and Department of Computing, Imperial College London, London, UK \and The Alan Turing Institute, London, UK \\
\email{\{saad.bashir,shan.raza,n.m.rajpoot\}@warwick.ac.uk, t.qaiser@imperial.ac.uk}}

\maketitle              % typeset the header of the contribution
\begin{abstract}
Semi-supervised techniques have removed the barriers of large scale labelled set by exploiting unlabelled data to improve the performance of a model. In this paper, we propose a semi-supervised deep multi-task classification and localization approach HydraMix-Net in the field of medical imagining where labelling is time consuming and costly. Firstly, the pseudo labels are generated using the model's prediction on the augmented set of unlabelled image with averaging. The high entropy predictions are further sharpened to reduced the entropy and are then mixed with the labelled set for training. The model is trained in multi-task learning manner with noise tolerant joint loss for classification localization and achieves better performance when given limited data in contrast to a simple deep model. On DLBCL data it achieves 80\% accuracy in contrast to simple CNN achieving 70\% accuracy when given only 100 labelled examples.

% \keywords{Semi-Supervised Learning \and Medical Imaging\and Deep Learning #and Cell Detection and Classification}
\end{abstract}
\section{Introduction}

Deep learning (DL) has revolutionized computer vision in recent years and achieved state-of-the-art performance in various vision-related tasks. The inevitable fact is that most of the DL success is attributed to availability of large scale datasets and compute-power available these days. To achieve state-of-the-art performance, it is incumbent to train models as single-task learning paradigm on large scale datasets with their associated labels. The costs associated with labelling of the datasets is often very high especially for medical imaging data which involves expert knowledge to collect the ground-truth. In contrast, semi-supervised learning (SSL) approaches \cite{semi_supervised} take advantage of the limited labelled data and leverages readily available unlabelled data to improve the model performance. This also alleviates the need for time-consuming and laborious task of manual annotations and assist training of more complex models for better performance. Generally, SSL techniques follow a two-step approach a) predict pseudo labels for unlabelled data from the model trained on limited labelled data and b) retrain the model on pseudo labels and limited labelled data to improve the performance. More recently, the trend has been to improve learning ability of SSL by introducing regularization \cite{augmentations,mixup} and entropy minimization \cite{entropy_min} to avoid high-density predictions and train models into an end-to-end manner.
\phantom{
with the help of model trained on small/limited labelled data and retrain the networks with the pseudo labelled and labelled data to improve the performance \cite{pseudo_label,pseudo_label_1}, where this small labelled data can be easily obtained from the experts in less time. Recently, the SSL techniques have forces on improving the loss terms by adding different learning techniques e.g. regularization \cite{augmentations,mixup} where the input is transformed differently to improve models learning by outputting the same labels. Entropy minimization \cite{entropy_min} were the high density predictions are avoided, Noise handling \cite{sce} where noisy labels are suppressed to improve the performance etc.}
In this work, we propose a multi-task SSL method to alleviate the need of time-consuming and laborious task(s) of manual labelling for histology whole-slide images (WSI). In this regard, we opted to use diffuse large B-cell lymphoma (DLBCL) data because manual annotation of cell type and nuclei localization is very hard  due to large number of cells present in WSIs. DLBCL malignancy originates from B-cell lymphocytes and it is the most common high-grade lymphoma among the western population with poor disease prognosis \cite{CHOP}. We propose a novel deep multi-task learning framework, HydraMix-Net, for simultaneous detection and classification of cells, enabling end-to-end learning in a semi-supervised manner. We improve the performance of a semi-supervised approach by enhancing a single loss term with noisy labels for joint training of multi-task problem which to our knowledge has not been performed earlier. Our main contributions are as follows: a) a novel multi-task SSL framework (HydraMix-Net) for cell detection and classification, and b) combating noisy labels using symmetric cross-entropy loss function.
% \phantom{
% \begin{itemize}
%   \item Proposing a novel semi-supervised approach HydraMix-Net for end-to-end learning of a multi-task training problem.
%   \item Improving the results with the help on combating the noisy labels in the training batches.
%   \item Patient wise extensive validation of the proposed approach.
% \end{itemize}
% }

\section{Related Work}
The purpose of semi-supervised task is to learn from unlabelled data during learning such that it improves the model's performance. To achieve this goal these approaches take advantage of different techniques to mitigate the issues faced during learning e.g., consistency regularization, entropy minimization and noise reduction etc. Decision boundary passing through high-density regions can be minimized using entropy minimization techniques like \cite{entropy_min} which minimize entropy with the help of a loss function for the unlabelled data. Consistency regularization can be achieved using standard augmentation such that the network knows if the input was being altered in some ways e.g., rotation, etc \cite{augmentations,mixup}. Semi-supervised approaches also suffer from noisy labels as the pseudo labels can introduce noise in the training batches wich can be handled using noise reduction methods such as \cite{sce}. Using these common approaches there have been semi-supervised methods for classification of natural images e.g., Berthelot et al. \cite{mixmatch} used simple data augmentation and mixup \cite{mixup} for consistency regularization and used sharping \cite{sharpen} for entropy minimization for semi-supervised training. Tarvainen et al. \cite{meanteacher} improved the temporal ensembling over labels to use moving average of the weights of student model in teacher model after comparing students prediction with its teacher's prediction, which in turn improves learning of the teacher model.  Inspired from all these methods and techniques we propose our novel deep multi-task join training framework for end-to-end classification and detection. Related work regarding fully supervised cell detection and classification is discussed in the Supplementary Material section 1.
 
\section{HydraMix-Net: Cell Detection and Classification}

\begin{figure}[t]
\begin{center}
\includegraphics[width=\textwidth]{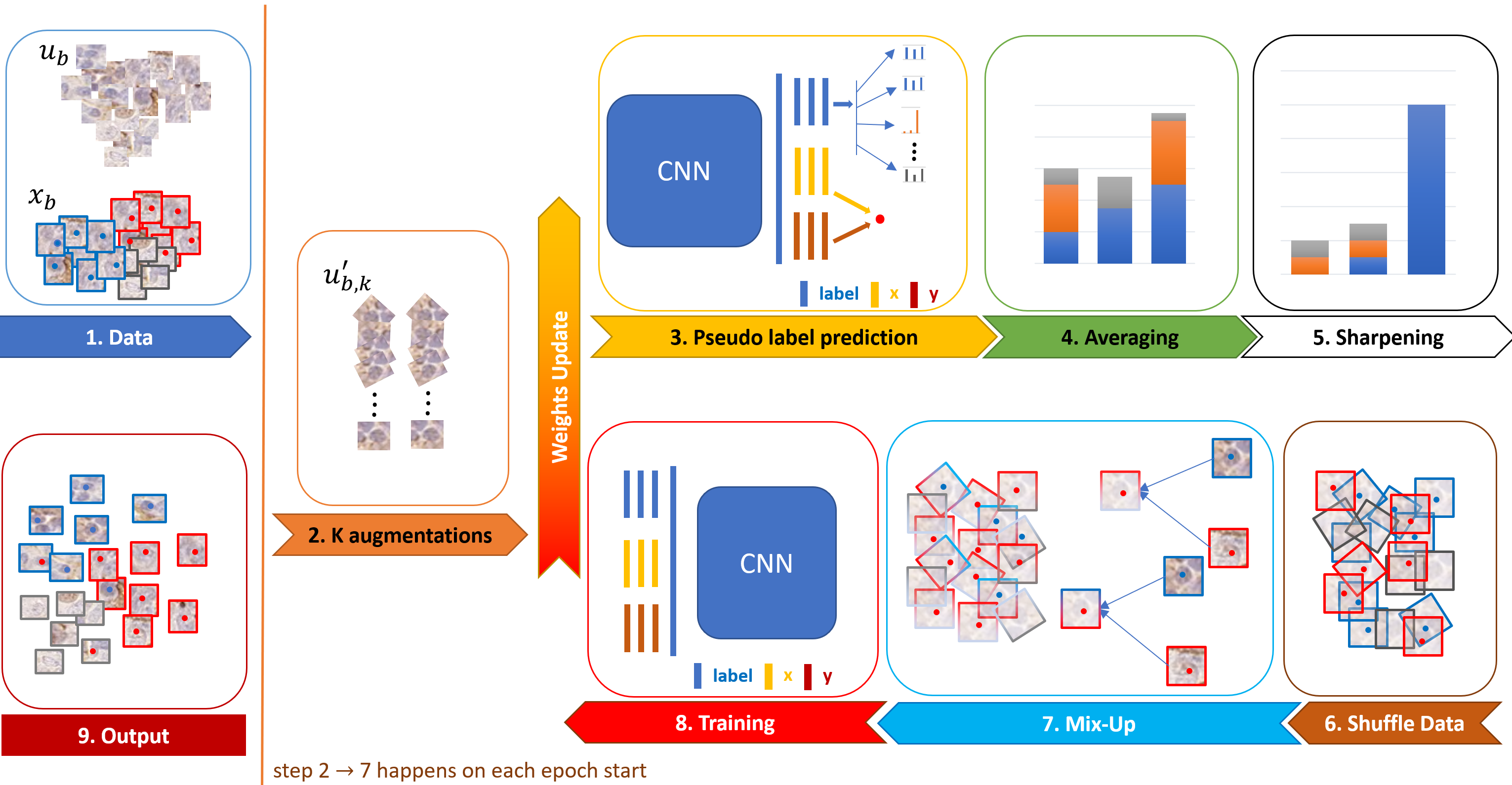}
\caption{The schematic diagram of the proposed HydraMix-Net. The unlabelled data $u_b$ is first subjected to \textit{k} augmentations to generate $u'_{b,k}$ and then process them from the model to generate pseudo labels after which the predictions are averaged and sharpened to minimize entropy in the prediction distribution. Once pseudo labels are assigned, unlabelled set $u_b$ is mixed-up with labelled data $x_b$ to help model iteratively learn more generalized distributions with noise suppression. }
\label{fig:main_model}
\end{center}
\end{figure}

The proposed semi-supervised method HydraMix-Net is a holistic approach consisting of different multi-task and semi-supervised techniques to handle various learning issues e.g., consistency regularization using standard augmentations and mixup techniques \cite{mixup}, entropy minimization with the help of sharpening \cite{sharpen}, and handling noisy labels with modified loss terms like symmetric cross entropy (SCE) loss \cite{sce}. The proposed HydraMix-Net jointly optimizes the combined loss function for classification and localization of centroids for the cell patches. Our proposed multi-task learning framework consists of a backbone model with three heads responsible for the classification and regression (i.e., localization of cell nuclei). The following sections delineate the data augmentation, pseudo label generation, noise handling and training in the proposed semi-supervised HydraMix-Net model, The schematic diagram of the proposed model can be seen in the Fig \ref{fig:main_model}.

\subsection{Data Augmentation}
 During training the model takes an input batch of labelled $x_b$ images from $X = \{ {x_b}\} _{b = 1}^B$ and unlabelled $u_b$ images from $U = \{ {u_b}\} _{b = 1}^B$, where $B$ was the total number of batches, with known one-hot encoded labels $l_c$ and $l_{x}, l_{y}$ representing nuclei centroid. To generate the pseudo labels and their centroids $l_{uc}, l_{ux},l_{uy}$ using the model, $k$ augmentations like horizontal flip, vertical flip, random rotate, etc., were applied to $u_b$ to yield an augmented batch $u'_b$ as ${u'_{b,k}} = augment(k,{u_b}),k \in (1,..,K)$. $x_b$ is also subject to single augmentation per image such that it generates $x'_b$ as ${x'_b} = augment(k,{x_b}),k = 1$.

\subsection{Pseudo Label Generation}
 To generate pseudo labels $l_{uc}$ for the batch $u_b$, predictions from the models $\varphi$ for $k$ augmented images $u'_b$ were averaged out on class distributions. While for pseudo centroids, prediction on only the original image from the model was used. This is due to the fact that after various augmentations, the centroids are not in the same place because of transformations and hence averaging the centroids of augmentations will lead of incorrect centroids as in eq. \eqref{eqt:pseudo}. 
 
 % ------------------------------ equation# 3 --------------------------%
 \begin{align}
{l_{uc}},{l_x},{l_y} = \left\{ {\begin{array}{*{25}{l}}
{\frac{1}{k}\sum\limits_{k = 1}^N {\varphi (y'|{u_{b,k}};\theta )} ,}&{\text{if } \;\; c  =  }1\\
{\varphi (y'|{u_b};\theta ),}&{{\rm{otherwise}}}
\end{array}} \right.
\label{eqt:pseudo}
\end{align}
  % ------------------------------ equation# 3 --------------------------%
\\
where $\varphi$ is the model and $\theta$ are the corresponding weights yielding the prediction $y'$ which was split into patch label $l_uc$ when $c$ = 1, otherwise centroids $l_{ux}$ and $l_{uy}$.
\\
\textbf{Pseudo Label Sharpening} The generated pseudo labels $l_{uc}$ tend to have large entropy in the prediction as a result of averaging of different distributions. Therefore, sharpening \cite{sharpen} was used to reduce or minimize entropy of predictions by adjusting temperature of the categorical distribution as in eq. \eqref{eqt:sharpe}.  
 
 % ------------------------------ equation# 4 --------------------------%
 \begin{align}
sharpening{(l_{uc},T)_i}: = {{\mathop {{l_i}}\limits^{\frac{1}{T}} } \mathord{\left/
 {\vphantom {{\mathop {{l_i}}\limits^{\frac{1}{T}} } {\sum\limits_j^L {\mathop {{l_j}}\limits^{\frac{1}{T}} } }}} \right.
 \kern-\nulldelimiterspace} {\sum\limits_j^L {\mathop {{l_j}}\limits^{\frac{1}{T}} } }}
\label{eqt:sharpe}
 \end{align}
  % ------------------------------ equation# 4 --------------------------%
where $l_{uc}$ is the categorical distribution of predictions averaged over $k$ augmentations and $T$ temperature is the hyper-parameter which controls the output distribution. When $T$ approaches to $0$ it will produce the one-hot encoded output meaning lowering the temperature will yield in low entropy output distributions. 

\subsection{Mixup}
To bridge the gap between unseen examples and remove over-fitting and achieve generalization in semi-supervised approaches mix-up \cite{mixup} technique was used. Given a pair of images and their labels as $(x_1,l_1)$ and $(x_2,l_2)$. Images were mixed along with their one-hot encoded labels in an appropriate proportion $\gamma$. However, the centroids were not mixed due to their numeric nature and transformations. Therefore, centroids from $x_1$ were used after fusion as shown in \eqref{eqt:mixup}. In our method, we have used the modified mix-up \cite{mixmatch} technique where $\gamma$ was extracted from beta distribution and then max between $\gamma$ and $1-\gamma$ was taken as $\gamma$, this ensures that maximum of the original image was preserved and output was closer to $x_1$.

 % ------------------------------ equation# 5 --------------------------%
 \begin{equation}
\begin{array}{l}
\gamma  = \max (Beta(\alpha ,\beta ) ,1 - Beta(\alpha ,\beta ) )\\
{x_m} = \gamma {x_1} + (1 - \gamma ){x_2}\\
{l_m} = \gamma {l_1} + (1 - \gamma ){l_2}\\
{l_{mx}}, {l_{my}} = {l_{x_1}},  {l_{y_1}}\\
\end{array}
\label{eqt:mixup}
 \end{equation}
  % ------------------------------ equation# 5 --------------------------%
In order to apply this technique here $x'_b$ and $u'_b$ were concatenated and shuffled into $W$ and were used for the mix-up. Afterwards, $x'_b$ was mixed-up with $W_{0...|x'_b|}$ and $u'_b$ was mixed-up with $W_{|x'_b|....N}$ where $|x'_b|$ is the length of the augmented mixed-up set $x'_b$ and N is the total number of samples in $W$.
 
 \subsection{Noise Reduction}
To handle noise, symmetric cross entropy (SCE) loss \cite{sce} was used for both labelled and unlabelled loss instead of just relying on categorical cross-entropy for labelled loss and mean squared loss for the guessed labels. SCE handles the noisy labels by incorporating cross-entropy term for labelled loss as well reverse cross-entropy for predictions loss. This provides a way to learn from model predictions as well instead of just relying on given labels as in eq. \eqref{eqt:sce}. As with iterative progressive learning, the model gets more confident in it's learning and predictions, which is why for unlabelled loss more weight is assigned to predictions and in labelled loss more weight is assigned to labels.
  % ------------------------------ equation# 5 --------------------------%
 \begin{equation}
{l_{sl}} = \delta (- \sum\limits_{c = 1}^C {q(c|{x_m})\log p(c|{x_m})}) + \rho (- \sum\limits_{c = 1}^C {p(c|{x_m})\log q(c|{x_m})})
\label{eqt:sce}
 \end{equation}
where $\delta$ and $\rho$ controls the effect of input labels and models predictions.
  % ------------------------------ equation# 5 --------------------------%
  
% \begin{figure}[t]
%   \includegraphics[width=0.9\linewidth ]{model_description.png}
%   \caption{(a) A schematic illustration of our proposed Hydra-Net and (b) description of different sub-components.}
%   \label{fig:boat1}
% \end{figure}

\subsection{Model Training:}
The learning mechanism of the HydraMix-Net jointly optimizes the combined loss function for classification and regression to predict label and location tuple for labelled and unlabelled batches as in eq.\eqref{eqt:loss}.

\begin{equation}
{l_{total}} = \mu ({l_{c - sce}} + {l_{uc - sce}}) + (1 - \mu )({l_{rx}} + {l_{ryx}} + {l_{ruy}} + {l_{ry}})
\label{eqt:loss}
\end{equation}
where $l_{c - sce}$ represents the symmetric cross-entropy loss for the labelled part where $l_{uc - sce}$ represents the symmetric cross-entropy loss for the unlabelled part, both coupled together in weight $\mu$ which weights the classification head more to provide more accurate labels. While the $l_{rx}$ and $l_{ry}$ are the mean squared error loss terms for the labelled data whereas the $l_{rux}$ and $l_{ruy}$ are the mean squared error loss terms for the unlabelled data for the regression head being weighted by the $(1-\mu)$.
While calculating loss for regression heads the predictions of the classification head were multiplied by regression heads in order to avoid the loss incorporated by background patches which is why the classification head was given more weight in the loss term.

\section{Results}
The data set used for the study is a private data for DLBCL \cite{tqiaser}. Patches of size 41 $\times$ 41 were extracted from 10 manually annotated WSI's resulting in 12553 patches and after offline augmentations, 24000 patches were used for this study. 3 WSI's were selected for the test purposes while 7 WSI's were used for the training purposes, splitting on 70-30 basis which resulted in 18000 training patches and 6000 test patches.\textbf{ See Supplementary Material section for the detailed description of the data set, implementation details, comparative and ablation study.}

\subsection{Experimental Settings}
The experimental settings used to test the effectiveness of the proposed approach were i) fully supervised ii) partial data iii) semi-supervised, In the first one all of the available data was used to train a simple CNN i.e., WideRes-Net \cite{wideresnet}, while in partial setting WideRes-Net was trained on partially labelled data. Lastly, HydraMix-Net used semi-supervised approach for training where both labelled and unlabelled data were used in a way discussed earlier in the section 3. Further, for labelled and unlabelled data we tested different configurations from 50 labelled images to 100, 200, 300, 500, 700 and so on.

% \subsection{Metrics}
% Standard matrices \cite{ref7} were used for the evaluation purposes i.e., Accuracy and F1 score for the experiments.

% % \begin{equation}
% % \text{Precision}_{~i} = \cfrac{M_{ii}}{\sum_j M_{ji}}
% % \label{eq:5}
% % \end{equation}
% % \begin{equation}
% % \text{Recall}_{~i} = \cfrac{M_{ii}}{\sum_j M_{ij}}
% % \label{eq:6}
% % \end{equation}
% \begin{equation}
% f1 = \frac{2 * (precision * recall)}{(precision + recall)}
% \label{eq:6}
% \end{equation}

% where ${M_{ii}}$ represents the current class predictions, and ${M_{ji}}$ is the prediction where all other classes are declared as $i$ in $j$, similarly the ${M_{ij}}$ is the prediction of $j$ in class $i$.

\subsection{Quantitative Results}
\begin{figure}[t]
\begin{center}
\includegraphics[width=\textwidth]{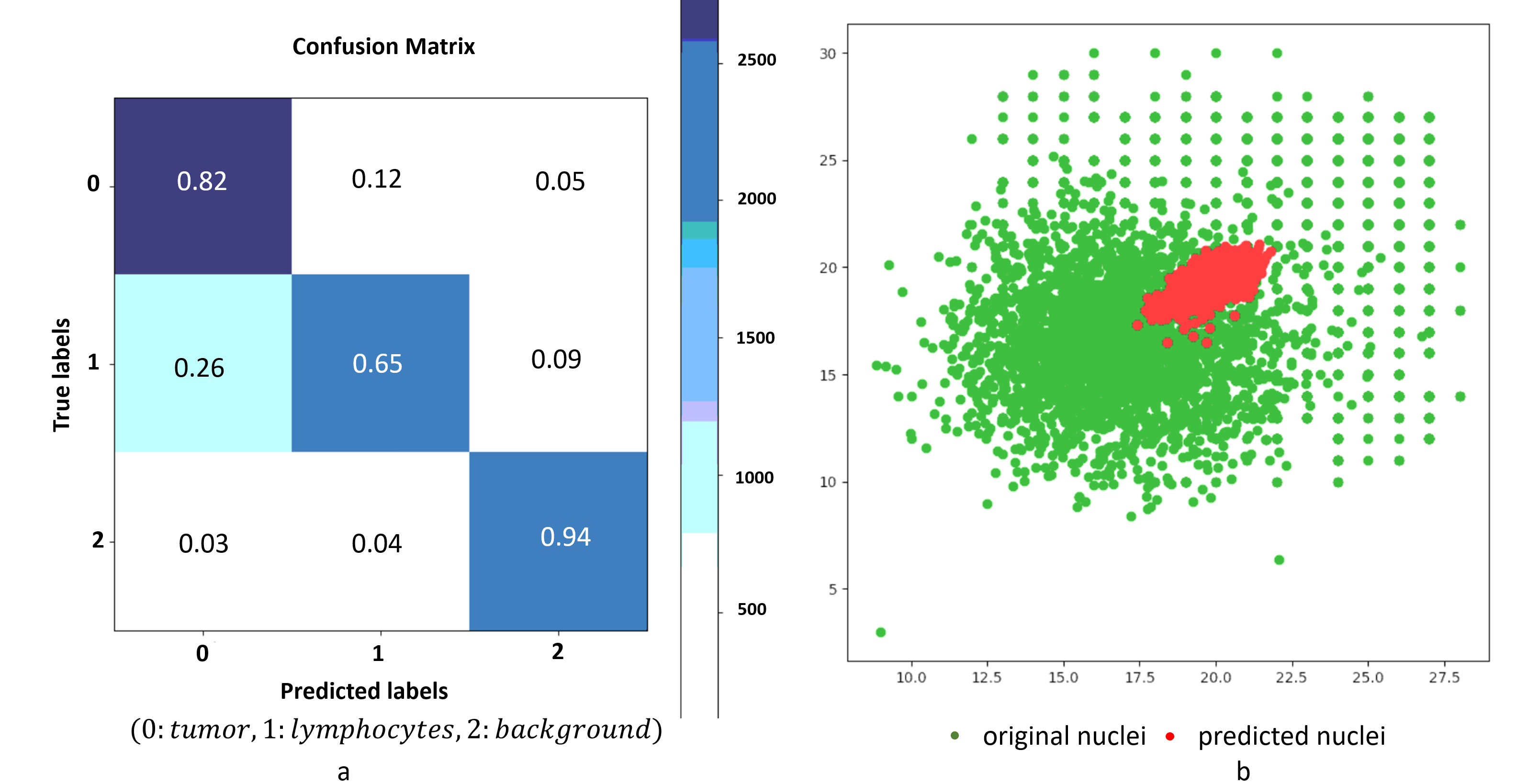}
\caption{(a) Represents the confusion matrix for the HydraMix-Net while (b) Represents the prediction and distribution of the centroid in the HydraMix-Net trained on 100 labelled instances where the output size is 32$\times$32. }
\label{fig:conf_nu}
\end{center}
\end{figure}

\begin{figure}[t]
\begin{center}
\includegraphics[width=\textwidth]{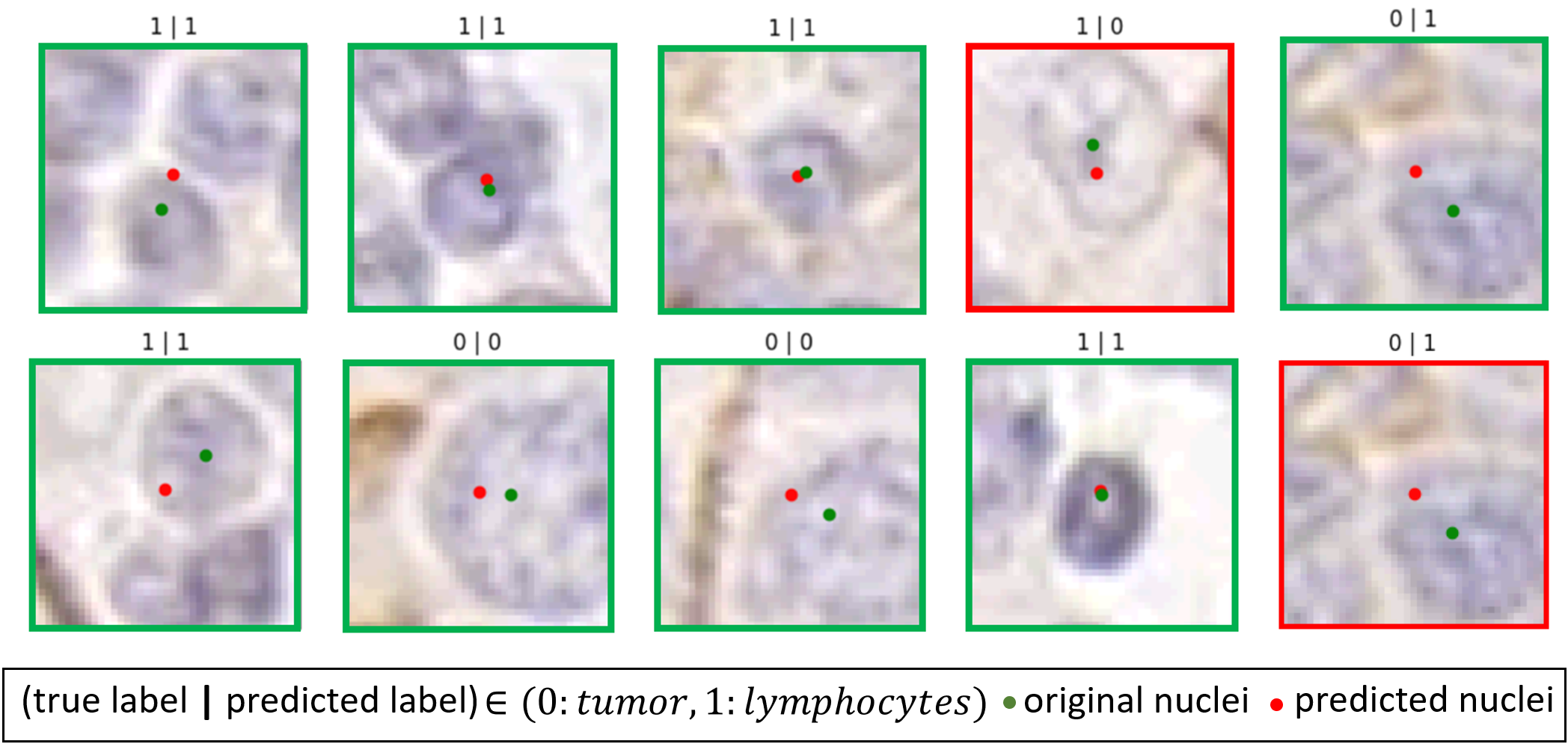}
\caption{The prediction of labels and distribution of the centroid on a example set where the HydraMix-Net was trained on 100 labelled examples }
\label{fig:100_pred}
\end{center}
\end{figure}

Table \ref{tabel:acc} shows the accuracy achieved by the HydraMix-Net in contrast to the simple CNN on partially labelled data e.g., when provided with the random 50 labelled examples the simple CNN model under-performed by achieving 62\% accuracy where the proposed approach leveraged the unlabelled data and achieved superior performance with 66\% accuracy. Similarly, when increased the data from 50 labelled examples to 100 and 300 the HydraMix-Net achieved higher performance and reached up to 81\% accuracy while simple CNN model trained on only these labelled examples only gave the best performance of 76\% accuracy which shows higher efficiency of the proposed approach in scarcity of the labelled examples. Confusion matrix for 100 labelled examples is shown along with the cell centroid detection in the Fig. \ref{fig:conf_nu}. Fig. \ref{fig:100_pred} shows the actual predictions for the proposed approach for the 100 labelled training set. When trained with all the data the highest accuracy achieved is 90\% where this threshold is reached by approx. 3000 labelled data by both the techniques.
\begin{table}[t]
\caption{Test accuracy of the HydraMix-Net and partial data approaches with various amount of labelled data provided. }
\centering
\begin{tabular*}{\textwidth}{@{\extracolsep{\fill}}| c | c | c | c | c | c | c | c |}
\hline
 labelled data & 50 & 100 & 300 & 500 & 700 & 1000 & 3000 \\
\hline
 \textbf{Simple CNN} & 0.62 & 0.70 & 0.76 & 0.83 & 0.85 & 0.84 & \textbf{0.90}\\ 
\hline
 \textbf{HydraMix-Net w/o SCE} & 0.66 & 0.70 & 0.70 & 0.35 & 0.35 & 0.35 & .-- \\ 
\hline
\textbf{HydraMix-Net} & \textbf{0.66} & \textbf{0.80} & \textbf{0.81} & \textbf{0.85} & \textbf{0.85} & \textbf{0.85} & 0.88 \\ 
\hline
\end{tabular*}
\label{tabel:acc}
\end{table}

% represents the confusion matrix for the simple CNN model trained on the partial data @100. It can be seen from the matrix that the false positives in the HydraMix-Net are less than the false positives in the partial data.

\section{Conclusion}
In this study, we proposed a novel end-to-end holistic multi-task SSL approach for simultaneous classification and localization of nuclei in DLBCL. Further, we plan to extend this work by improving the technique with the help of strong augmentations and validating the performance of our HydraMix-Net on larger cohorts from multiple tumour indications. The cell detection and classification may also be help in performing follow-up analysis like survival prediction and understanding the spatial arrangement of malignant cells within tumour micro-environment to predict other clinical outcomes.


\bibliographystyle{splncs04}

%%%%%%%%%% Merge with supplemental materials %%%%%%%%%%
\pagebreak
\begin{center}
\textbf{\large Supplemental Materials: HydraMix-Net: A Deep Multi-task Semi-supervised Learning Approach for Cell Detection and Classification}
\end{center}

\section{Related Section}
\subsection{Cell Classification and Detection}
In terms of cell classification and detection, Cirean et al.\cite{Cirean} proposed a simple deep learning based classification model to differentiate between the mitotic and non-mitotic cells in the breast WSI's. Sirinukunwattana et al.\cite{Korsuk} used the locality sensitive information for the localization of the cell nuclei while used the Neighboring Ensemble Predictor (NEP) for the classification purposes. Qaiser et al.\cite{tqiaser} proposed the joint multi-task framework to explored the spatial arrangements of the tumour cells and their localisation with the collagen VI in DLBCL by proposing the novel digital proximity signature (DPS) marker in the tumour rich collagen regions.

\section{Dataset}
32 WSI's were collected for this study stained with the immunohistochemistry and Hematoxylin counter-stain to simultaneously detect collagen VI and nuclear morphology. The cohort included 10 samples from females and 2 from male for the DLBCL where the age ranging from 24 years to 90 years old. The ground truth for cell detection and classification was done for 10 cases by an expert pathologist in VSM tool and 2617 cells were annotated in total where 2039 were tumours cells, 462 lymphocytes and 116 macrophages. Patches of size 41 $\times$ 41 were extracted resulting in 12553 patches and due to inherent class imbalance in patches, offline augmentations including flipping, rotation and crop were applied to balance the dataset resulting in 24000 patches of equal distribution of 8000 patches for each class. 3 WSI's were selected for the test purposes while 7 WSI's were used for training purposes, splitting on 70-30 basis which resulted in 18000 training patches and 6 thousand test patches.

\section{Results}
\subsection{Implementation Details}
The proposed approach was implemented in TensorFlow 2.0 where the base CNN was selected as WidesResNet \cite{wideresnet} with an additional three heads i) classification head ii) two-regression heads. In the classification head, the final output of the WideResNet was global average pooled and passed through three dense layers of sizes 128, 64 and 32 before the classification layer while in the regression heads takes the flatten layer results of output layer which is then passed through 2 dense layers of sizes 128 and 32 before going to the regression output. The dense layers were activated using the ReLu activation with \textit{l}2 regularization. The model was optimized with the Adam optimizer with the adaptive learning rate from 0.001 to 0.00001 trained for 100 epochs and batch size of 32.

\subsection{Results}
\begin{figure}[!]
\begin{center}
\includegraphics[width=\textwidth]{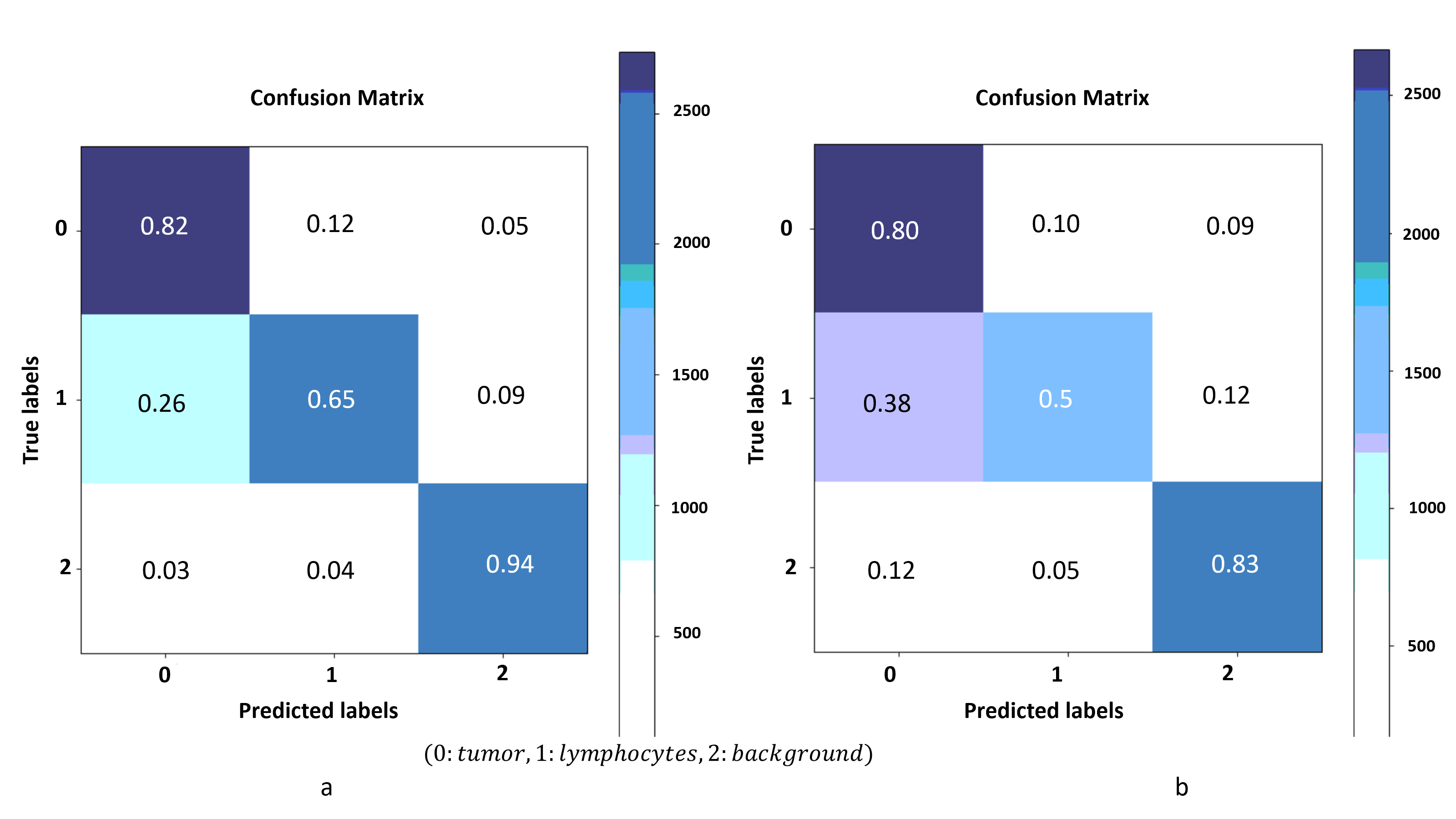}
\caption{(a) Represents confusion matrix for the HydraMix-Net while (b) represents confusion matrix for simple CNN model trained on partial data of size 100. It can be seen from matrix that false positives in the HydraMix-Net are less than false positives in partial data. }
\label{fig:conf100}
\end{center}
\end{figure}

\begin{figure}[!]
\begin{center}
\includegraphics[width=\textwidth]{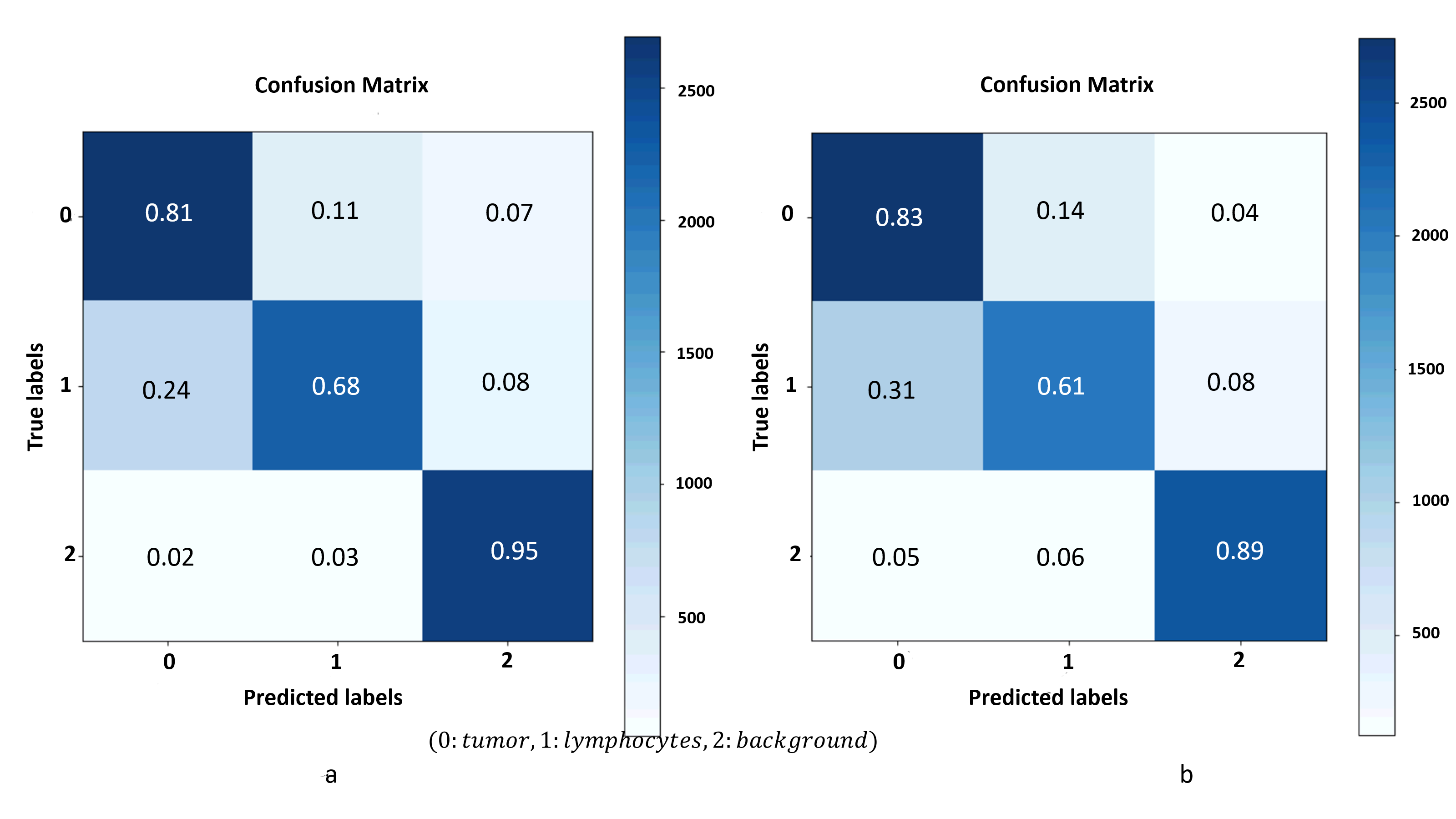}
\caption{(a) Represents confusion matrix for the HydraMix-Net while (b) represents confusion matrix for simple CNN model trained on partial data  of size 300. It can be seen from matrix that false positives in the HydraMix-Net are more in case of tumour while for background and lymphocytes false positives in partial data training are in abundance. }
\label{fig:conf300}
\end{center}
\end{figure}

\begin{figure}[!]
\begin{center}
\includegraphics[width=\textwidth]{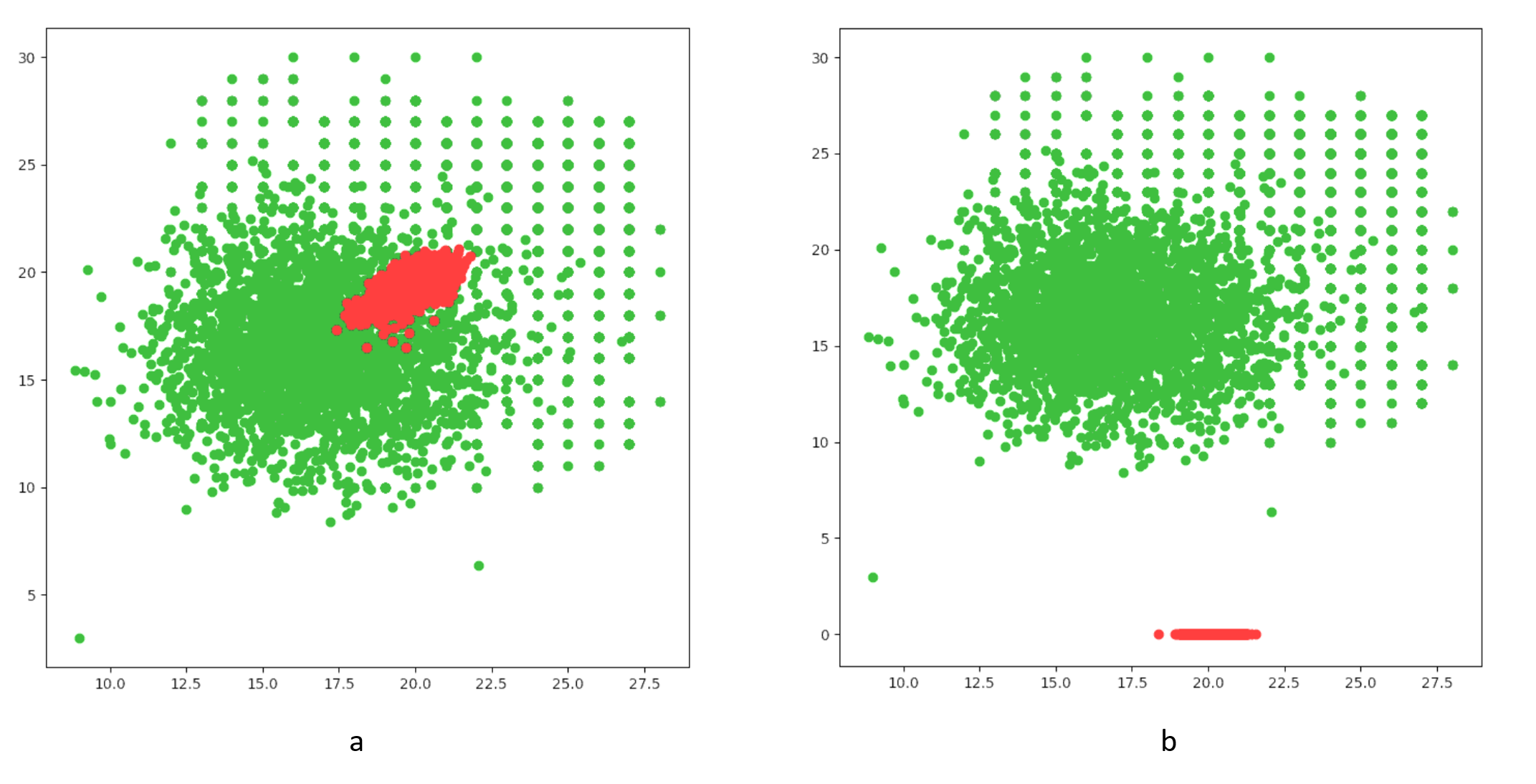}
\caption{(a) Represents prediction and distribution of centroid in the HydraMix-Net trained while (b) shows distribution of centroid learned by simple model on partial data of size 100. It can be seen that simple model fails to learn the approximate distribution in one axis. however, the HydraMix-Net can learn with the help on unlabelled data.}
\label{fig:nu100}
\end{center}
\end{figure}

\begin{figure}[!]
\begin{center}
\includegraphics[width=\textwidth]{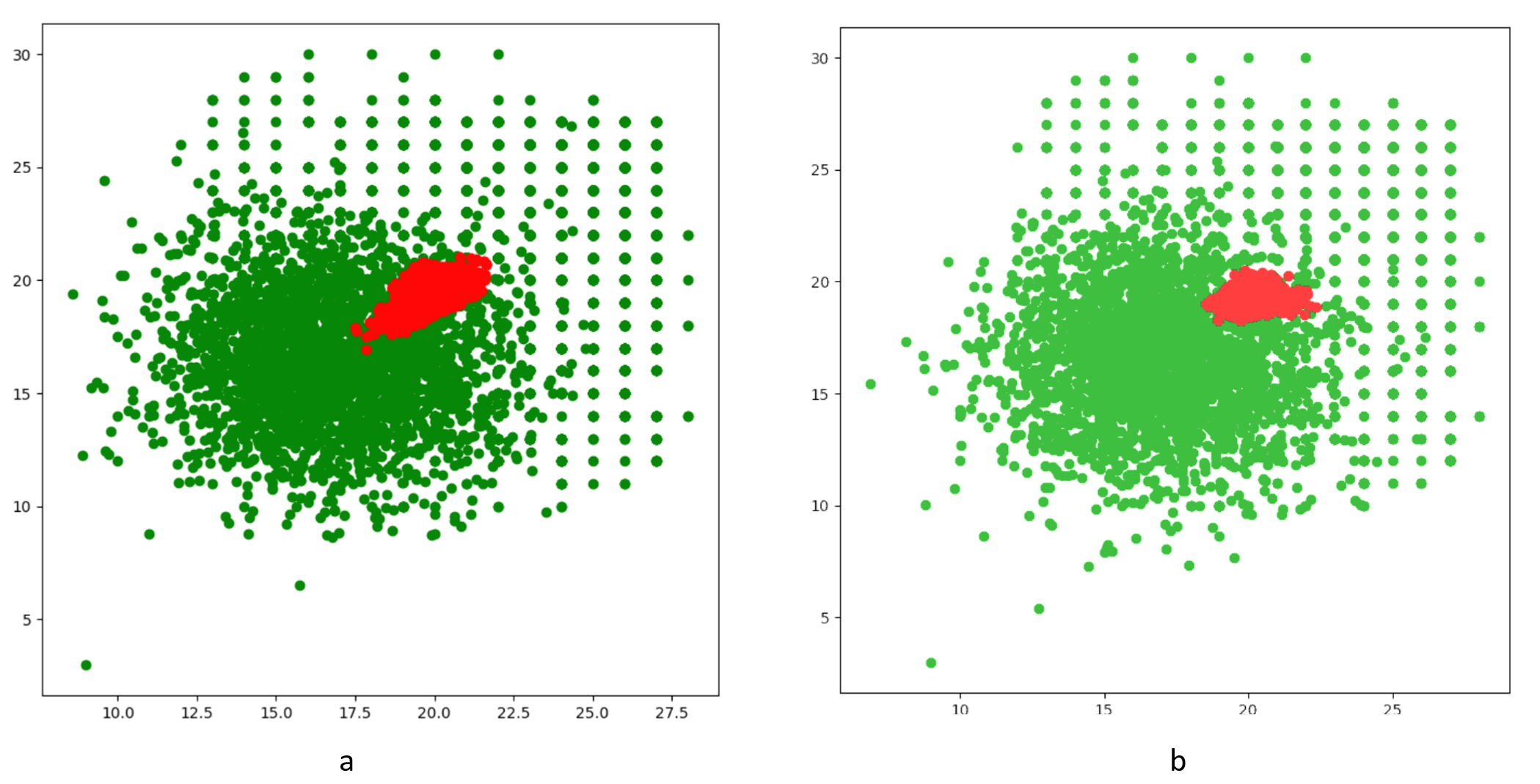}
\caption{(a) Represents prediction and distribution of centroid in the HydraMix-Net trained while (b) shows distribution of centroid learned by simple model on partial data of size 300. It can be seen that simple model's centroid distribution is less sparse and is more compact while HydraMix-Net is trying to learn the sparse distribution.}
\label{fig:nu300}
\end{center}
\end{figure}

\begin{figure}[t]
\begin{center}
\includegraphics[width=\textwidth]{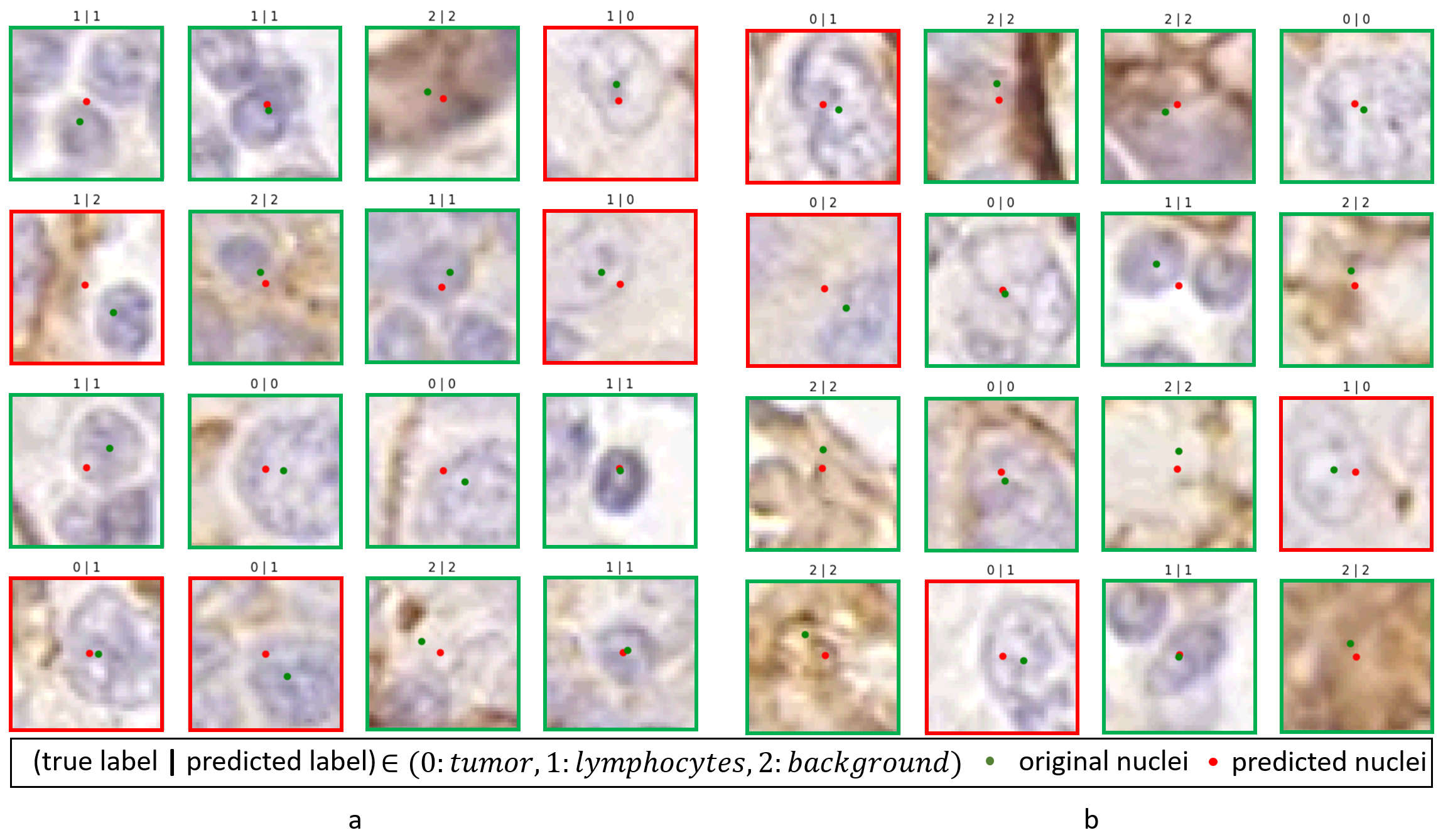}
\caption{(a) Shows prediction and distribution of the centroid in the HydraMix-Net trained on 100 labelled examples (b) shows prediction and distribution of the centroid in the HydraMix-Net trained on 300 labelled examples. }
\label{fig:preds}
\end{center}
\end{figure}

 Fig.\ref{fig:conf100}, \ref{fig:conf300} shows comparative results of the proposed approach with simple CNN model on the 100 labelled set and 300 labelled set. While Fig. \ref{fig:nu100}, \ref{fig:nu300} shows results of nuclei distribution learned by the proposed model and simple CNN trained on 100 labelled set and 300 labelled set, where it can be seen that the simple models failed to learn the distribution in very limited data availability. Further, it can be seen that nuclei locations are biased towards the centre of the patch because of the inherent biasness in the training data. Fig. \ref{fig:preds} shows predictions for the proposed approach for the 100 labelled set. in (a) while 300 labelled set in (b) and it can be seen that the model is learning to classify the patch accurately along with nuclei prediction among tumour, lymphocytes and background patches.

\subsection{Discussion}

\subsubsection{Noise Reduction}
\begin{figure}[!]
\begin{center}
\includegraphics[width=\textwidth]{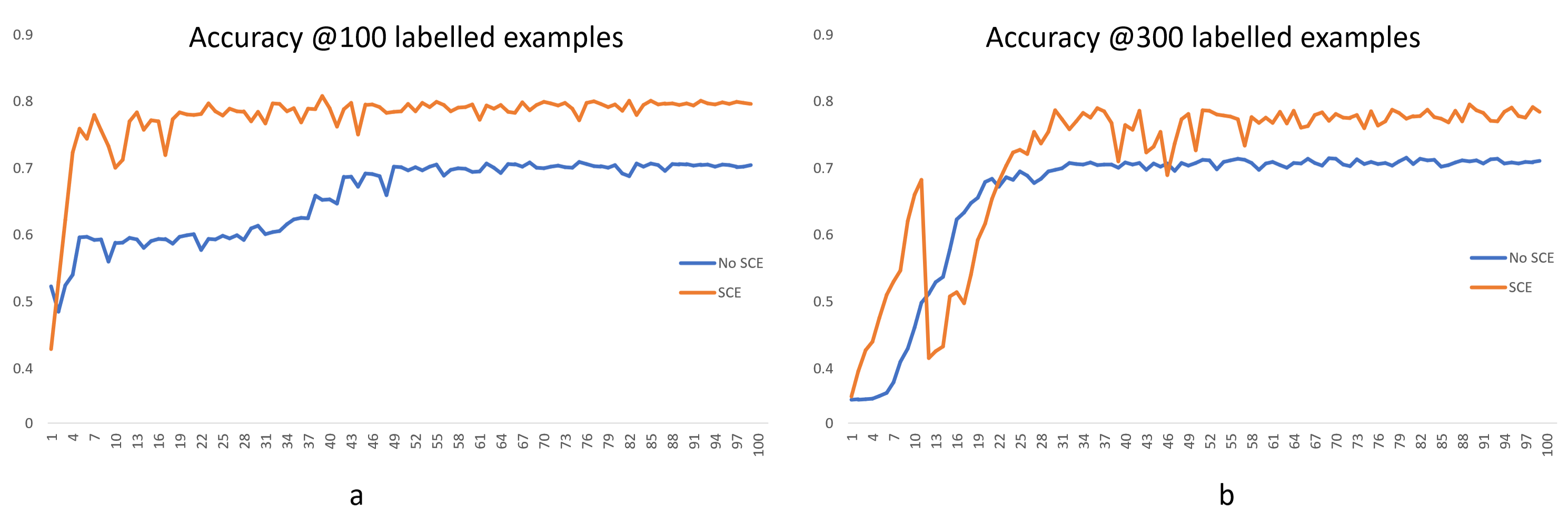}
\caption{(a) Represents the accuracy curves of the models trained with 100 labelled examples with the orange line showing the model with SCE and the blue line showing model without SCE and it can be seen that the model without SCE under-performs the model with SCE with a margin of 10\% in accuracy. Similarly (b) Represents the accuracy curves of the models trained with 300 labelled examples with the orange line showing the model with SCE and the blue line showing model without SCE and it can be seen that the model without SCE under-performs the model with SCE with a margin of 5\% in accuracy.}
\label{fig:sce_loss}
\end{center}
\end{figure}

In this study, we have included symmetric cross-entropy loss to reduce the effect of the noisy label and ease out learning capabilities. Labelled data was given more weightage while computing the SCE loss because there is less noise in the labelled set and labels are not much noisy (i.e., mixup doesn't add much noise in the labels) while in the case of unlabelled data loss the new predictions were given more importance as it was believed that the newly predicted values were more accurate as the model has learnet and improved the previous mistakes. Hence, we experimented with few configurations to see the effectiveness of SCE loss and turned out that addition of SCE made model learn more than simple cross-entropy and \textit{l}2 loss as it can be seen in the Fig  \ref{fig:sce_loss}. Interestingly, when more data is provided, the chances of model overfitting increases as training is sensitive towards the noise and start to overfit the dominant class distribution which is seen in Table \ref{tabel:acc}. Hence, adding SCE improves overall learning of the approach by making this technique less susceptible to noise.

\subsubsection{Knowledge vs Accuracy}
\begin{figure}[!]
\begin{center}
\includegraphics[width=\textwidth]{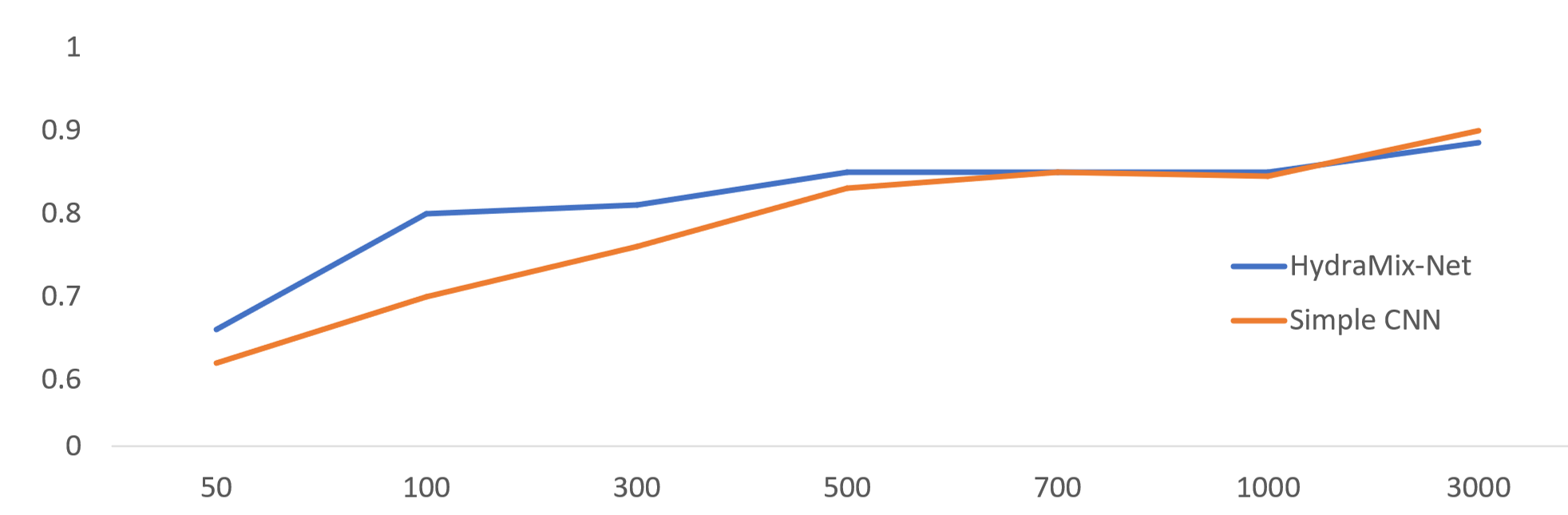}
\caption{Represents the increase in knowledge vs increase in accuracy where the knowledge is the number of labelled samples which can help the model to learn more accurately on the true labels and it can be seen that the proposed approach HydraMix-Net leverages semi-supervised approach and outperformed the simple CNN trained on partial data.}
\label{fig:inc_K_acc}
\end{center}
\end{figure}

In this study, we have also examined behaviour of increasing the knowledge i.e., increasing labelled samples while training corresponding to the model's accuracy it has been shown through experiments that increasing knowledge does increases the accuracy. As with more accurate labelled data training model gets chance to learn it more accurately and performs better on validation and test sets as it can be seen in the Table \ref{tabel:acc} and in Fig. \ref{fig:inc_K_acc}. 

\bibliographystyle{splncs04}

\end{document}